\documentclass[11pt]{article}
\usepackage[a4paper,margin=24mm]{geometry}
\usepackage[T1]{fontenc}
\usepackage[utf8]{inputenc}
\usepackage{lmodern}
\usepackage{amsmath,amssymb,amsfonts,amsthm,mathrsfs}
\usepackage{graphicx,float,booktabs}
\usepackage{microtype,placeins}
\usepackage[numbers,sort&compress]{natbib}
\usepackage{caption}
\usepackage[hidelinks]{hyperref}
\usepackage{bookmark}
\newcommand{\DMF}{\textsc{DMF-Gen}}
\newcommand{\bmhead}[1]{\par\medskip\noindent\textbf{#1}\enspace}
\title{Sparse Reconstruction by Direct Measurement-to-Field Generation in Function Space}
\author{Linzheng Wang$^{1*}$, Jason Chen$^1$, Nicolas Tricard$^1$, Zituo Chen$^1$,\\
Xingsen Guo$^{2,3}$, Sili Deng$^{1*}$\\[0.6em]
\small $^1$Department of Mechanical Engineering, Massachusetts Institute of Technology,\\
\small Cambridge, MA 02139, USA\\
\small $^2$Department of Civil, Environmental and Geomatic Engineering, University College London,\\
\small London WC1E 6BT, UK\\
\small $^3$Department of Engineering, University of Cambridge, Cambridge CB2 1PZ, UK\\[0.4em]
\small $^*$Correspondence: \texttt{wanglz@mit.edu}; \texttt{silideng@mit.edu}}
\date{}
\begin{document}
\maketitle
\begin{abstract}
Understanding physical systems often requires recovering their full spatial state from measurements available only at sparse locations. This reconstruction becomes challenging when physical boundaries are unknown, high-resolution training fields are unavailable, or important quantities cannot be measured directly. 
Sparse observations may also be consistent with multiple plausible states, leading to significant reconstruction uncertainty. 
Here we introduce Direct Measurement-to-Field Generation (\DMF{}), a functional rectified flow framework that learns how sparse measurements drive physical-field generation end to end. 
Instead of relying on grid-based rasterization or runtime guidance, \DMF{} jointly learns a sensor--global--sensor representation with the generative transport, dynamically fusing local grounding with global context to evaluate physical fields at arbitrary query coordinates.
Across benchmarks involving unobserved boundaries, missing high-resolution training fields, and unmeasured physical quantities, \DMF{} outperforms alternative baselines by significantly reducing reconstruction errors and preserving the fidelity of physical properties. 
Furthermore, its stochastic ensemble generation reliably quantifies uncertainty while improving continuous ranked probability scores.
These results extend sparse reconstruction from fixed-setting point estimation toward measurement-conditioned generation of physical fields under incomplete and changing information.
\end{abstract}

\noindent Sparse reconstruction seeks to recover a spatially distributed physical state from measurements available at only a small number of locations~\cite{Manohar2018SparseSensor,Fukami2021Voronoi, Santos2023Senseiver}.
This need arises in many applications, from flow diagnostics~\cite{Callaham2019Sparse} and reacting-flow imaging~\cite{Wang2024ChemFlow} to environmental sensing~\cite{Gou2024Global}, geophysical monitoring~\cite{Lenain2026unprecedented} and data-driven digital twins~\cite{Wang2026DigitalTwin}.
Because the measurements constrain only a small fraction of the unknown field, the inverse problem is underdetermined.
Many fields agree with the same observations, so a reconstruction method must supply the missing structure from prior knowledge of which physical states are admissible.
Approaches to supplying this structure have progressed from reduced-order estimation~\cite{Willcox2006Gappy} and sparse sensor placement optimization~\cite{Brunton2016Placement} to learned reconstruction of high-dimensional and multiphysics states~\cite{Achermann2024WindSeer,Li2024S3GM,Wang2026UnifiedManifold}.

As these methods move toward practical use, the missing information extends beyond the gaps between sensors in a fixed field.
The number, placement, and type of sensors can change between experiments or deployments~\cite{Li2024S3GM,Liu2025Enhancing}.
The physical domain itself can be unknown, as when the shape of a defect or an immersed body is not observed directly~\cite{Li2023GeoFNO}.
The field may be needed at a finer resolution than the complete training data provide~\cite{Zhao2024RecFNO}.
In coupled systems, some quantities of interest may not have direct measurements at all~\cite{Wang2026Developing}.
Each of these conditions enlarges the set of physical states consistent with the observation evidence~\cite{Wright2009Sparse}.
A reconstruction method for such settings therefore faces two demands.
First, it must accept observations and produce outputs whose form adapts across different measurements and experiments.
Second, it must describe the variation that the measurements leave unresolved, rather than collapse it into a single prediction~\cite{Yao2025FunDPS,mohseni2025spectral}.

Existing reconstruction methods offer different strengths in meeting these requirements.
Reduced-order and sparse-representation methods exploit low-dimensional structure, so that a few measurements can constrain a compact set of basis coefficients~\cite{Callaham2019Sparse}.
Methods from image processing, commonly using convolutional networks as backbones, learn spatial structure directly from data, but they require irregular observations to be mapped onto structured inputs~\cite{Fukami2021Voronoi}.
Set- and attention-based models such as Senseiver remove this restriction by accepting variable sensor collections and predicting the field at requested coordinates~\cite{Santos2023Senseiver}.
Neural operators, in turn, learn mappings between functions that can be evaluated across discretizations and geometries~\cite{Lu2021DeepONet,Li2021FNO,Kovachki2023NeuralOperator,Azizzadenesheli2024NeuralOperators}.
In their standard form, however, these models are deterministic and return one field for each observation set.
When several states are consistent with the measurements, a single field cannot show how those states differ~\cite{Huang2024DiffusionPDE}.
Localized structures whose positions are uncertain, such as a thin flame front or a narrow stress band, tend to be averaged into smoothed features that match none of the plausible states.

Generative models have begun to serve this role in recent PDE inference and turbulence reconstruction works~\cite{Huang2024DiffusionPDE, Yao2025FunDPS,Du2024CoNFiLD,Wang2026FunDiff}.
These methods construct a sample by progressively transforming random noise into a physical field.
For sparse reconstruction, they differ mainly in two design choices: what object is generated, and how the measurements influence its generation.
The generated object may be a field sampled on a fixed grid~\cite{Li2024S3GM}, or a compressed latent code that a separately trained autoencoder or neural field decodes into the physical field~\cite{Rombach2022LDM,Du2024CoNFiLD,Wang2026FunDiff}.
Grid-based generation ties the result to one discretization.
Latent generation reduces cost, but its accuracy depends on how well the learned compression preserves fine-scale structure.
Work on direct generation in the original data space has further shown that such compression is not required even for high-dimensional data~\cite{Hoogeboom2023SimpleDiffusion,Chen_2026_CVPR,ICLR2026_47f8478e,Li_2026_CVPR}.
Function-space generative models avoid both limitations by treating the generated field as a function of spatial coordinates~\cite{Kerrigan2024FFM,Lim2025FunctionSpaceDiffusion,Zhang2026FRF}.
In flow-based versions of these models, a random function is gradually transformed along a learned velocity over an artificial generation time~\cite{Liu2023RectifiedFlow}.
A grid then serves only to specify where the resulting function is evaluated.
This formulation is a natural foundation for sparse reconstruction, as the observations and the requested outputs are both sets of coordinates in the same physical domain.

Less developed is how sparse measurements should drive function-space generation.
In many existing generators, measurements either guide sampling from a prior trained without them~\cite{Yao2025FunDPS}, or are supplied as value and mask maps on the grid of the generated field~\cite{Li2024S3GM}.
With guidance, agreement with the data is imposed during sampling, so the relationship between measurements and the unobserved field is not itself learned.
Grid-based conditioning, by contrast, ties the observations to a prescribed spatial discretization and channel layout, making irregular or changing sensor sets cumbersome to represent and often requiring rasterization or interpolation before reconstruction~\cite{Berner2026Principled}. 
Addressing both limitations calls for a conditioning representation that is learned jointly with the generative transport and operates on the measurement set itself. 
Such a representation should have three properties.
First, it should integrate nonlocal information across sensors and physical variables, since sparse reconstruction can depend on both long-range spatial correlations and cross-variable coupling~\cite{Calvello2025ContinuumAttention}.
Second, it should keep this information attached to the locations where it was acquired, since each observation constrains the field most strongly nearby.
Third, both the global context and the spatially anchored evidence should remain available when evaluating the generative velocity at arbitrary query coordinates throughout the transport.

\begin{figure}[htbp]
    \centering
    \includegraphics[width=\linewidth]{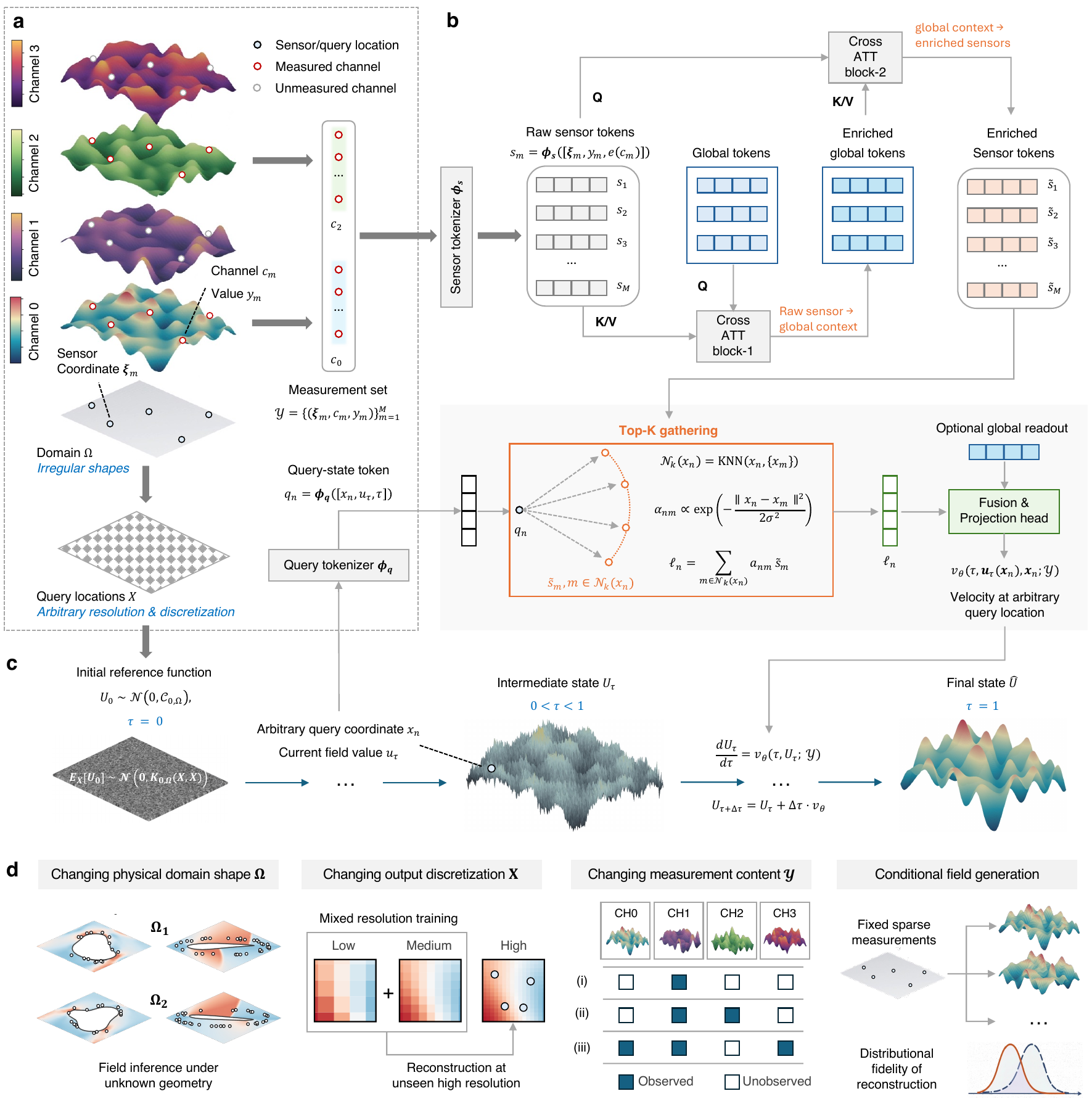}
    \caption{\textbf{Direct measurement-to-field generation.} \textbf{a}, Each observation records a coordinate, physical channel, and value. Query coordinates are specified separately. \textbf{b}, Sensor tokens exchange information through global latents and return context to sensor locations. Local evidence and a global readout condition the velocity at each query coordinate. \textbf{c}, Rectified flow transports a smooth Gaussian reference function to a conditional field. Independent draws give multiple reconstructions. \textbf{d}, Evaluations vary the domain $\Omega$, output coordinates $X$ and measurements $\mathcal{Y}$.}
    \label{fig:Overview}
\end{figure}

To meet these requirements, we introduce \DMF{}, which formulates sparse reconstruction as a direct measurement-conditioned transport of the physical field in function space (Fig.~\ref{fig:Overview}).
Its measurement representation is learned jointly with the transport, so the model learns how observations should shape generation instead of having agreement imposed only during sampling.
The representation encodes each observation by its coordinate, measured value, and physical-channel identity (Fig.~\ref{fig:Overview}a), such that sensor sets of any size, layout, or channel composition enter the model without rasterization onto a grid.
Adapted from global--local sensor reasoning~\cite{Wang2026GLU}, a sensor--global--sensor cycle first aggregates multi-variable cross-sensor information into a compact set of global latents and then broadcasts it back, enabling each spatially anchored token to integrate global context with its own measurement (Fig.~\ref{fig:Overview}b).
During generation, evaluating velocity at any requested coordinate leverages the nearest enriched sensor features for local evidence and a coordinate-dependent global readout for field-wide context (Fig.~\ref{fig:Overview}c).
Throughout this process, the evolving state is the physical field itself, rather than a compressed latent code or a fixed array.
Generation therefore proceeds directly from measurements to field without a separately trained field autoencoder, and the same model can produce reconstructions flexibly on a requested grid, mesh, or point set.

We evaluate \DMF{} against four needs that arise in scientific sparse reconstruction (Fig.~\ref{fig:Overview}d).
First, when the physical domain varies and its boundary is not observed, the field and the domain must be inferred together.
\DMF{} represents both on a common enclosing domain and generates the domain mask alongside the physical fields.
We test this on elasticity and airfoil benchmarks with hidden boundaries~\cite{Li2023GeoFNO}.
Second, high-resolution fields may be needed when complete training fields exist only at coarser resolutions.
Coordinate-wise transport allows training on mixed resolutions~\cite{Shi2025MINO}.
We test this on two-dimensional Navier--Stokes flow without complete high-resolution training fields~\cite{Takamoto2022PDEBench}.
Third, the measured variables may change, and some quantities of interest may remain unmeasured, so information must pass between physical fields.
Channel-labelled observations provide a common representation for different combinations of measured variables. 
We test this on turbulent combustion with unmeasured species, velocity and pressure fields~\cite{Wang2026GLU}.
Finally, when the measurements remain compatible with several physical states, repeated generation should describe this unresolved variation and indicate which reconstructions are least reliable.

\section*{Results}
\label{sec: Results}

\subsection*{Direct measurement-to-field generation in function space}
\label{Subsec: ModelOutline}

We consider a physical field with $C$ channels, such as temperature, velocity, and species mass fractions.
Each scalar observation records its coordinate, physical-channel identity, and value,
\begin{equation}
\mathcal{Y}
=\{(\boldsymbol{\xi}_m,c_m,y_m)\}_{m=1}^{M},
\qquad
y_m=U_{1,c_m}(\boldsymbol{\xi}_m)+\varepsilon_m.
\label{eq:results-measurements}
\end{equation}
The reconstruction is evaluated on a separate set of requested coordinates, $X=\{\boldsymbol{x}_n\}_{n=1}^{N}$, where the complete $C$-channel field is predicted.

The target distribution is $\nu_{\Omega}^{\mathcal{Y}}=\operatorname{Law}(U_1\mid\mathcal{Y})$, the law of the physical field conditioned on the measurements.
Following functional rectified flow~\cite{Zhang2026FRF}, we approximate this distribution by learning a transformation from a random reference field,
\begin{equation}
\frac{\mathrm{d}U_{\tau}}{\mathrm{d}\tau}
=V_{\theta}(\tau,U_{\tau};\mathcal{Y}),
\qquad
U_0\sim\mathcal{N}(0,C_{0,\Omega}),
\qquad 0\leq\tau\leq1.
\label{eq:results-transport}
\end{equation}
The flow time $\tau$ indexes generation, not physical evolution.

The central architectural requirement is to convert a finite measurement set into a velocity at every requested coordinate. We first tokenize each observation as
\begin{equation}
s_m=\phi_s\!\left(
[\gamma_s(\boldsymbol{\xi}_m),y_m,e(c_m)]
\right),
\label{eq:results-token}
\end{equation}
where $\gamma_s$ encodes coordinates and $e(c_m)$ identifies the measured variable.
A fixed-size global latent array reads the sensor tokens through cross-attention and integrates information across them.
The sensor tokens then read the processed global latents.
This feedback produces enriched tokens $\widetilde{s}_m$ that remain attached to their original coordinates.

At each requested coordinate, the velocity model combines the current field state with measurement information at both local and global scales.
A query-state token $q_n$ represents the coordinate $\boldsymbol{x}_n$, the current value $U_{\tau}(\boldsymbol{x}_n)$ and the flow time.

The $K$ nearest enriched sensor anchors provide local evidence,
\begin{equation}
\ell_n
=\sum_{m\in\mathcal{N}_K(\boldsymbol{x}_n)}
\alpha_{nm}\widetilde{s}_m,
\qquad
\alpha_{nm}
=\operatorname{softmax}_{m}\!\left[
-\frac{\|\boldsymbol{x}_n-\boldsymbol{\xi}_m\|_2^2}{2\sigma^2}
+b_{\theta}(\widetilde{s}_m)
\right],
\label{eq:results-local}
\end{equation}
where the learned bias adjusts the importance of sensors within the geometric neighborhood. A coordinate-based readout of the global latents provides complementary field-wide information $g_n$. The shared velocity head combines these terms,
\begin{equation}
\left[V_{\theta}(\tau,U_{\tau};\mathcal{Y})\right]
(\boldsymbol{x}_n)
=\psi_{\theta}(q_n,\ell_n,g_n).
\label{eq:results-velocity}
\end{equation}

For each target $U_1$, an independent reference draw $U_0$ and a uniform flow time define the straight interpolation $X_{\tau}=(1-\tau)U_0+\tau U_1$. The model predicts its displacement $U_1-U_0$ while receiving measurements sampled from $U_1$,
\begin{equation}
\mathcal{L}_{\mathrm{RF}}
=\mathbb{E}\!\left[
\frac{1}{NC}\sum_{n=1}^{N}
\left\|
v_{\theta}
(\tau,X_{\tau}(\boldsymbol{x}_n),\boldsymbol{x}_n;\mathcal{Y})
-\bigl(U_1(\boldsymbol{x}_n)-U_0(\boldsymbol{x}_n)\bigr)
\right\|_2^2
\right].
\label{eq:results-loss}
\end{equation}
This objective trains the measurement encoder and velocity network jointly without integrating a trajectory during optimization. At inference, repeated source draws produce an empirical conditional ensemble. 

We organize the baselines to separate three modeling choices in sparse reconstruction.
First, FFM-FNO~\cite{Kerrigan2024FFM} and FFM-Perceiver~\cite{Jaegle2022PerceiverIO} use the same function-space flow-matching objective with Fourier-operator and latent-attention parameterizations, respectively.
Second, SiT~\cite{Ma2024SiT} and Latent FM~\cite{Dao2023LatentFM} provide generative comparisons with different representations of the evolving state.
SiT transports patch-based physical-field arrays, whereas Latent FM transports an autoencoded field representation.
Third, we compare with deterministic sparse reconstructors.
These include supervised FNO~\cite{Li2023GeoFNO,Zhao2024RecFNO}, coordinate-query attention through Senseiver~\cite{Santos2023Senseiver}, and an MLP-RBF model that combines local radial-basis aggregation with global pooled context.
Within each evaluation setting, all methods receive the same prescribed observations and predict the same target fields.

\subsection*{Changing the domain: sparse inversion on irregular geometries}
\label{Subsec: IrregularGeo}

Sparse measurements can provide information about both the physical field and the domain on which that field is defined.
We test this joint field--geometry inversion using the elasticity and airfoil datasets released with Geo-FNO~\cite{Li2023GeoFNO}.
The elasticity cases contain stress fields around irregular holes representing different defect geometries, whereas the airfoil cases contain compressible flow around changing shapes.
The nominal observations are 16 von Mises stress measurements for elasticity and 32 static-pressure measurements for airfoil flow. The targets are the complete stress field or the airfoil density, velocity, and pressure fields, respectively, together with the unobserved physical-domain mask. The Mach number is derived from the reconstructed airfoil state to examine coherence across the inferred fields.

\begin{figure}[htbp!]
    \centering
    \includegraphics[
        width=\linewidth,
        height=0.70\textheight,
        keepaspectratio
    ]{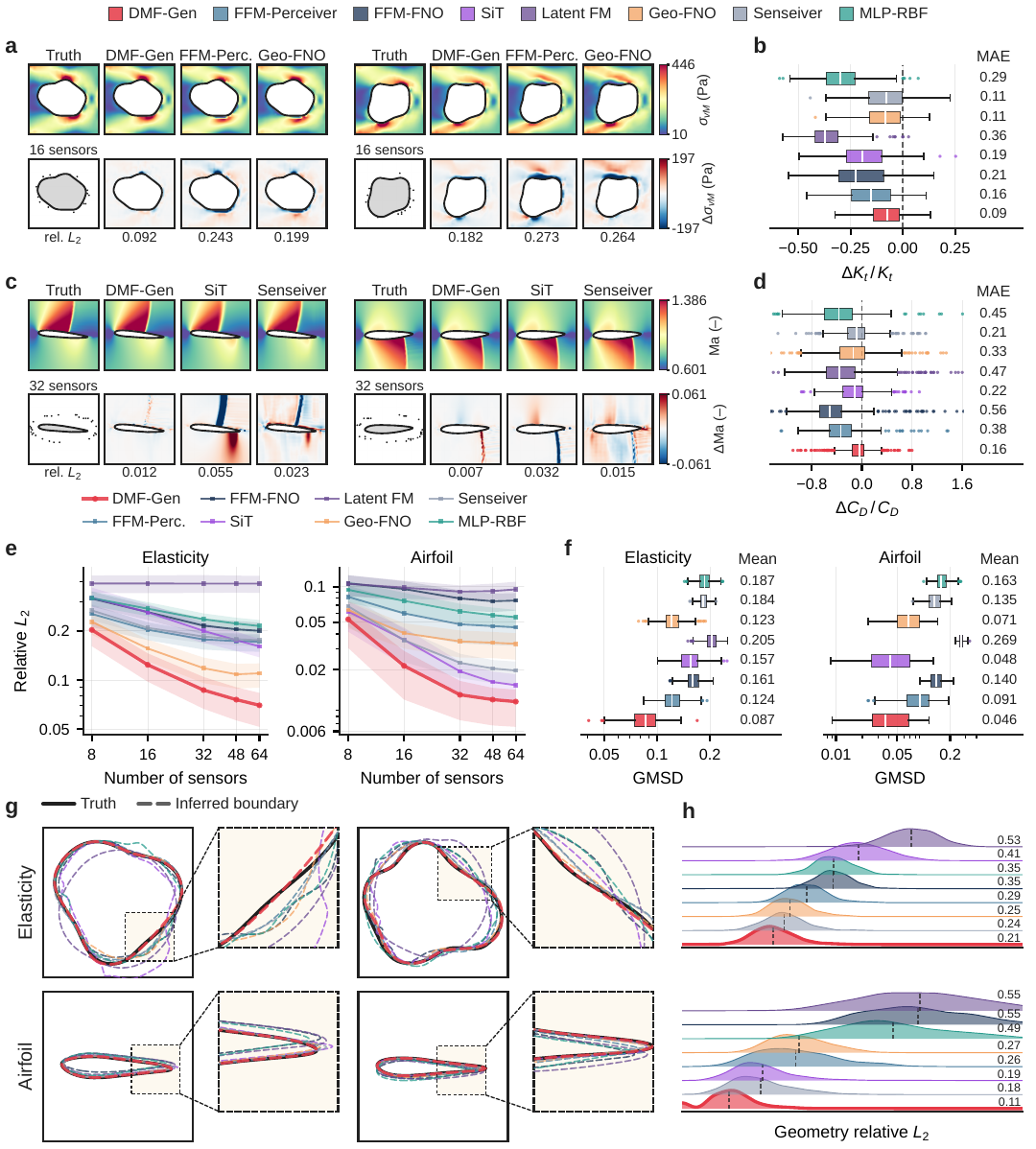}
    \caption{\textbf{Field and hidden-boundary reconstruction on irregular domains.} Results include 400 elasticity and 498 airfoil test cases. \textbf{a,c}, Von Mises stress and derived Mach number from 16 stress and 32 pressure measurements. \textbf{b,d}, Signed normalized errors in $K_t$ and $C_D$. Labels give mean absolute normalized errors. \textbf{e}, Mean masked relative-$L_2$ error across sensor counts, with interquartile bands. \textbf{f}, Gradient-magnitude similarity deviation (GMSD) distributions and means. \textbf{g}, True and inferred boundaries for selected cases. \textbf{h}, Thresholded geometry relative-$L_2$ error distributions and means. 
    }
    \label{fig:irr_geometry}
\end{figure}

The selected fields in Fig.~\ref{fig:irr_geometry}a,c illustrate the structures that dominate reconstruction difficulty. \DMF{} more closely localizes these structures in the illustrated cases. Across all 400 elasticity and 498 airfoil evaluation cases, its mean masked relative-$L_2$ errors were 0.1238 for stress and 0.0122 for Mach number, 20.5\% below Geo-FNO and 36.7\% below SiT, respectively. This metric measures the field discrepancy on the true physical domain, separating errors in reconstructed values from errors in the predicted mask.

The derived quantities in Fig.~\ref{fig:irr_geometry}b,d test whether improved field recovery carries over to physical analysis. The peak-to-mean stress-concentration proxy $K_t$ measures localized amplification relative to the mean material stress. \DMF{} achieved the lowest mean absolute normalized $K_t$ error, 0.0935, reducing the error by 13.1\% relative to Geo-FNO. The pressure-drag coefficient $C_D$ is computed by integrating the reconstructed static pressure over the true airfoil boundary. Its mean absolute normalized error was 0.1632, 22.4\% below Senseiver. 

The sensor-count sweep tests whether reconstruction remains effective as the available number of measurements changes.
\DMF{} achieved the lowest mean field error at every tested count from 8 to 64 sensors, including counts below and above the training ranges of 16--32 for elasticity and 32--48 for airfoil flow. 

Narrow stress bands around the hole and sharp Mach-number transitions around the airfoil can be smoothed or displaced even when global field error is low. We therefore evaluated gradient-magnitude similarity deviation (GMSD) to assess the preservation of local gradient structure~\cite{Xue2014GMSD} (Fig.~\ref{fig:irr_geometry}f). For elasticity, \DMF{} achieved the lowest mean GMSD of 0.087 compared with 0.123 for Geo-FNO, supporting better recovery of the localized stress bands illustrated in Fig.~\ref{fig:irr_geometry}a. For airfoil flow, \DMF{} had the lowest mean GMSD at 0.046. 

Geometry recovery further evaluates the unobserved domain mask separately from field accuracy. Elasticity sensors are selected from a sample-dependent band near the hole, so their coordinates contain partial boundary information. Airfoil sensors lie on the same enclosing ellipse for every profile, requiring the boundary to be inferred from pressure values. 
Across the full evaluation sets (Fig.~\ref{fig:irr_geometry}h), \DMF{} had the lowest mean geometry relative-$L_2$ errors with approximately 0.21 for elasticity and 0.11 for airfoil flow. Senseiver, the best baseline in both datasets, had corresponding errors of 0.24 and 0.18.

\subsection*{Changing the output discretization: reconstruction without complete high-resolution training fields}
\label{Subsec: MixedResolution}

Sparse reconstruction may require a field to be recovered at a finer spatial resolution than is available in the complete fields used for training.
This situation can arise when high-resolution full-field data are expensive or scarce, while sparse measurements remain available for the state to be reconstructed~\cite{Zhao2024RecFNO}.

\begin{figure}[htbp]
    \centering
    \includegraphics[width=\linewidth]{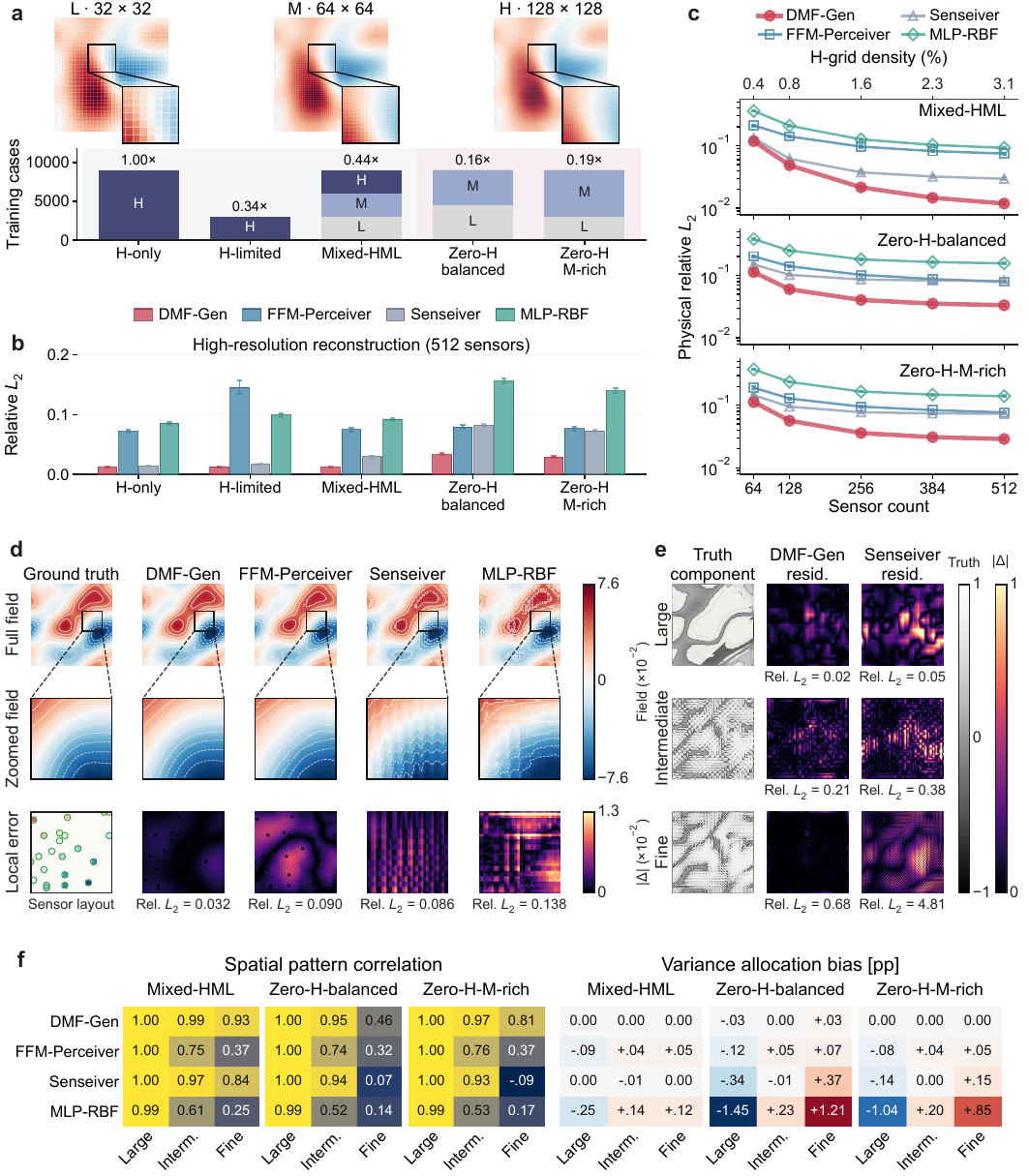}
    \caption{\textbf{Reconstruction across training and output resolutions.} 
    \textbf{a}, Low- (L, $32\times32$), medium- (M, $64\times64$) and high-resolution (H, $128\times128$) fields and five training recipes. 
    Bar labels give spatial-degree-of-freedom exposure relative to H-only.
    \textbf{b}, Mean H-resolution physical relative-$L_2$ error using 512 sparse sensors. Whiskers are 95\% simple case-bootstrap intervals. 
    \textbf{c}, Error across 64--512 nested H-grid sensors. 
    \textbf{d}, Full field, matched enlargement and local absolute error for a held-out Zero-H-M-rich state at 512 sensors. 
    \textbf{e}, Wavelet-scale truth and absolute residuals for \DMF{} and Senseiver on the same state at 256 sensors. 
    \textbf{f}, Median scale-wise pattern correlation and squared-energy allocation bias over 300 held-out case--time pairs at 256 sensors.}
    \label{fig:mixed_resolution}
\end{figure}

The 2-D Navier--Stokes data from PDEBench~\cite{Takamoto2022PDEBench} provide time-dependent fields on a common coordinate domain. 
We selected a scalar field and represented it on low- (L, $32\times32$), medium- (M, $64\times64$) and high-resolution (H, $128\times128$) grids (Fig.~\ref{fig:mixed_resolution}a).
All models are evaluated on held-out H-resolution fields. At inference, all models receive only sparse point measurements and reconstruct the field on the H-resolution output coordinates.
Five training recipes progressively change how much high-resolution information is available to the model (Fig.~\ref{fig:mixed_resolution}a). H-only contains complete H-resolution training fields. H-limited reduces their number. Mixed-HML combines L-, M-, and H-resolution fields within one training recipe. Zero-H-balanced and Zero-H-M-rich contain no complete H-resolution training fields. Zero-H-M-rich includes more M-resolution data. 

We compare \DMF{}, FFM-Perceiver, Senseiver and MLP-RBF.
This comparison evaluates reconstruction across training-resolution recipes on the same H-resolution output grid.
At 512 sensors, \DMF{} achieved the lowest mean H-resolution error under all five training recipes (Fig.~\ref{fig:mixed_resolution}b). Its mean relative-$L_2$ errors were $0.0121$, $0.0129$ and $0.0119$ under H-only, H-limited and Mixed-HML training, respectively, showing that its H-resolution accuracy was retained when lower-resolution fields were mixed with H-resolution training data.
With no complete H-resolution training fields, \DMF{} had mean errors of $0.0334$ for Zero-H-balanced and $0.0288$ for Zero-H-M-rich. These were 57.8\% and 60.5\% lower than the strongest evaluated alternative under the corresponding recipes. 

The sensor sweeps in Fig.~\ref{fig:mixed_resolution}c distinguish information learned from the training fields from additional sparse evidence available at deployment. 
Across Mixed-HML and both zero-H recipes, \DMF{} achieved the lowest mean error at every tested sensor count from 64 to 512. 
Thus, even when H-resolution fields are absent from training, sparse measurements continue to constrain the reconstructed high-resolution state.

We further tested whether this advantage persists across spatial scales using a four-level wavelet decomposition (Fig.~\ref{fig:mixed_resolution}e).  Across 300 fields (Fig.~\ref{fig:mixed_resolution}f), all methods recover the large-scale component well. Their main differences appear at intermediate and fine scales. Under Zero-H-M-rich training, \DMF{} retained median spatial pattern correlations of $1.000$, $0.969$ and $0.815$ across the large, intermediate and fine components. The corresponding fine-scale correlations were $0.368$ for FFM-Perceiver, $-0.089$ for Senseiver, and $0.165$ for MLP-RBF. 
\DMF{} also kept the median squared-energy allocation bias within 0.005 percentage points of the reference at every scale. 

\subsection*{Changing the measurement content: reconstruction of unobserved fields}
\label{Subsec: MultiField}

In many sensing problems, some physical quantities are difficult or unavailable to measure directly, while other coupled variables can be observed more readily~\cite{Wang2026UnifiedManifold}.
For sparse reconstruction, the challenge becomes more demanding when entire target fields contain no direct observations~\cite{Gomes2025sparse}.
We therefore change which physical variables are measured while keeping the domain, query coordinates, and full target state fixed~\cite{Kobayashi2026VirtualSensing}.
We used an in-house OpenFOAM simulation of methane--air turbulent combustion over a two-dimensional backward-facing step~\cite{Wang2026GLU}. The dataset contains methane and carbon monoxide mass fractions, temperature, streamwise velocity, and pressure, $\{Y_{\mathrm{CH_4}},Y_{\mathrm{CO}},T,U_1,p\}$, at 40,300 spatial locations. 

\begin{figure}[htbp]
    \centering
    \includegraphics[width=0.95\textwidth]{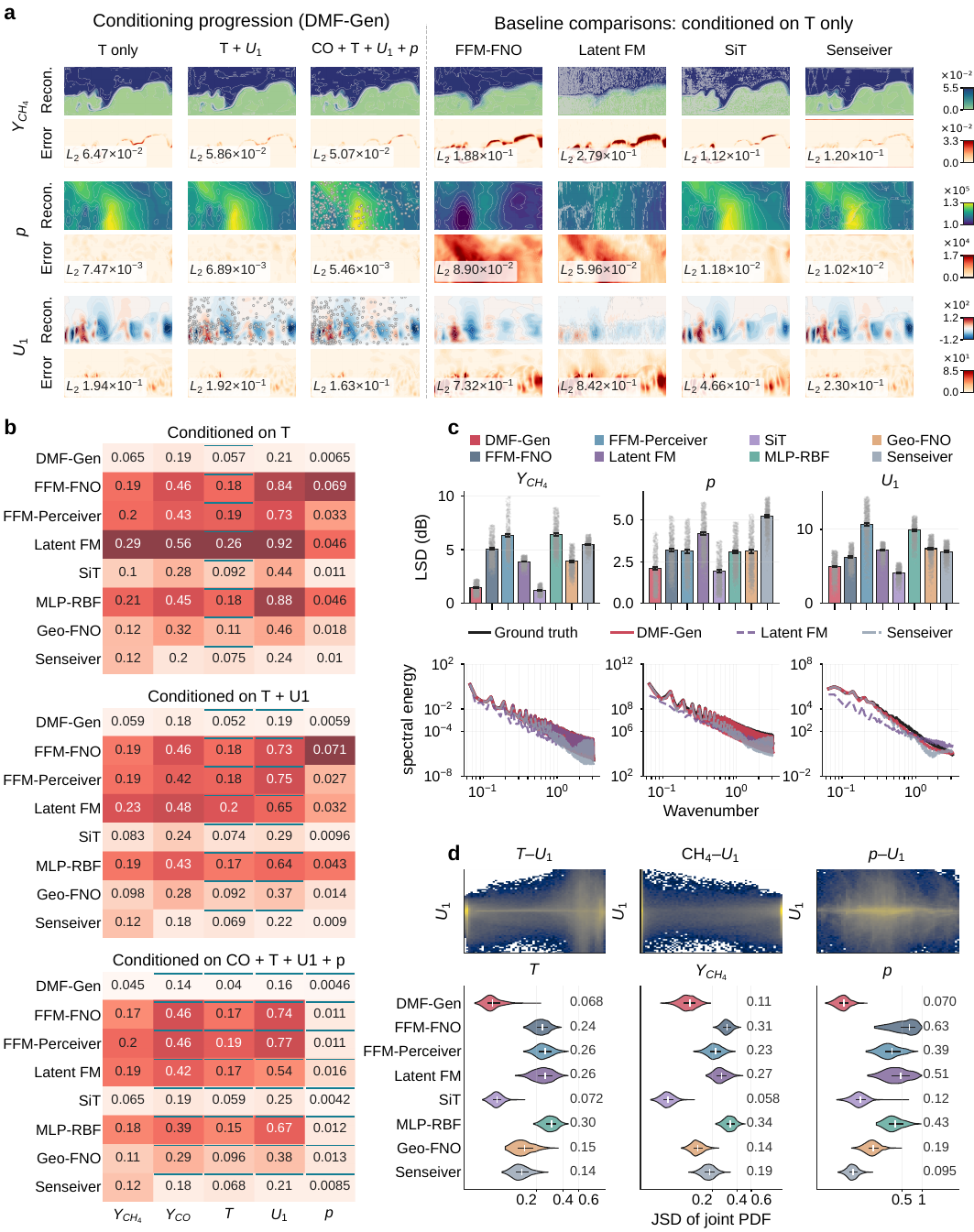}
    \caption{\textbf{Reconstruction of missing channels in turbulent combustion.} 
    \textbf{a}, \DMF{} reconstructions and absolute errors under three observed-channel sets with 256 sensors per channel. 
    Selected temperature-only baselines appear on the right. 
    \textbf{b}, Mean physical relative-$L_2$ error by field over 1,000 evaluation states. Teal marks observed fields. \textbf{c}, Spectra and mean log-spectral distance for unobserved $Y_{\mathrm{CH_4}}$, $p$ and $U_1$ under temperature-only conditioning. Error bars show 95\% simple snapshot-bootstrap intervals. 
    \textbf{d}, Ground-truth joint PDFs pooled over 25 states and per-state base-2 Jensen--Shannon divergence (JSD) distributions over 1,000 states for $T$--$U_1$, $Y_{\mathrm{CH_4}}$--$U_1$ and $p$--$U_1$. 
    }
    \label{fig:MultiFieldReconstruction}
\end{figure}

We used three observation regimes. The first contains 256 $T$ measurements. The second contains 256 measurements each of $T$ and $U_1$. The third contains 256 measurements each of $Y_{\mathrm{CO}}$, $T$, $U_1$ and $p$. Every method reconstructed all five fields. The regimes contain 256, 512 and 1,024 scalar observations, respectively, so broader channel coverage also increases the total amount of evidence. $Y_{\mathrm{CH_4}}$ remains unobserved in all three regimes. 

Across 1,000 test states, \DMF{} had the lowest mean error over the channels absent from each observation regime. Its unobserved-channel means were $0.117$, $0.0802$ and $0.0454$ for temperature-only, temperature--velocity and four-channel conditioning, respectively. These values were approximately 18\%, 22\% and 31\% lower than the strongest alternative under the corresponding regime (Senseiver for the first two and SiT for the third). 

We assessed spatial-scale content separately for the three displayed unobserved fields under temperature-only conditioning. As shown in Fig.~\ref{fig:MultiFieldReconstruction}c, for the unobserved $Y_{\mathrm{CH_4}}$, $p$ and $U_1$ fields, \DMF{} obtained mean log-spectral distances of $1.431$, $2.092$ and $4.966$\,dB, respectively, ranking second for all three. SiT had the lowest corresponding means, $1.239$, $1.926$ and $4.121$\,dB.

Spatial spectra describe each field separately, while coupled-field statistics test relationships between reconstructed variables (Fig.~\ref{fig:MultiFieldReconstruction}d). 
Under temperature-only conditioning, \DMF{} had the lowest mean base-2 Jensen--Shannon divergence for the thermal--flow pair $T$--$U_1$ ($0.068$) and for the pressure--velocity pair $p$--$U_1$ ($0.070$). The latter is a strict missing-channel test because neither variable was observed. SiT had the lowest divergence for the chemistry--flow pair $Y_{\mathrm{CH_4}}$--$U_1$ at $0.058$, compared with $0.11$ for \DMF{}. 
Both variables were unobserved.

\subsection*{Conditional field generation and reconstruction mechanisms}
\label{subsec: EnsembleMechanismCost}

\begin{figure*}[htbp]
    \centering
    \includegraphics[width=\textwidth]{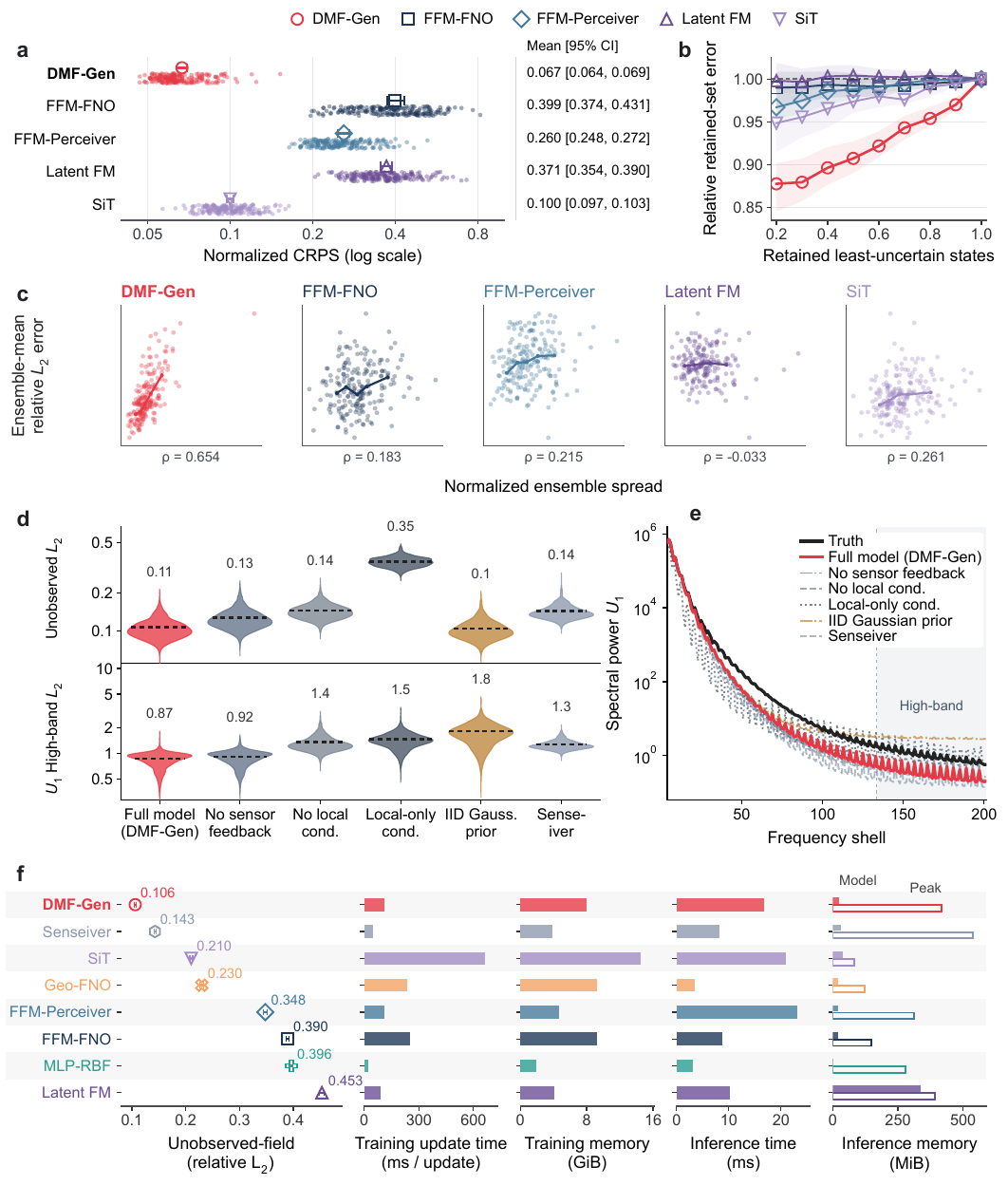}
    \caption{\textbf{Conditional generation, ablations, and cost.} All analyses use 256 $T$ measurements and target the four unobserved combustion fields. 
    \textbf{a}, Normalized pointwise continuous ranked probability score (CRPS) from 64 draws for each of 200 paired states. 
    \textbf{b}, Reconstruction error after retaining states with the lowest ensemble spread, normalized by each method's full-cohort mean. Whiskers in a and bands in b are 95\% moving-block-bootstrap intervals. 
    \textbf{c}, Ensemble spread versus ensemble-mean relative-$L_2$ error for the same states. Labels give Spearman $\rho$. Facet axes differ. 
    \textbf{d,e}, Aggregate and high-band errors and median $U_1$ spectra from 1,000-state ablation comparisons. 
    \textbf{f}, Reconstruction error, time per training update, inference latency, and GPU memory use. 
    }
    \label{fig:uq_cost}
\end{figure*}

For each of 200 evaluation states under temperature-only conditioning, we generated 64 conditional samples with independent reference-function draws while holding the same 256 $T$ measurements fixed. 
The continuous ranked probability score (CRPS) jointly reflects agreement with the realized target and dispersion within the predictive ensemble, with lower values indicating better distributional reconstruction. 
Across the 200 states, \DMF{} achieved a mean normalized CRPS of 0.0667 (95\% CI, 0.0640--0.0694), compared with 0.0999 (0.0970--0.1030) for the next-lowest method, SiT (Fig.~\ref{fig:uq_cost}a), corresponding to a 33.2\% reduction.

We then tested whether ensemble spread could be used to preferentially retain easier reconstruction cases. Within each method, states were ranked from lowest to highest ensemble spread, and the mean reconstruction error of the retained subset was normalized by that method's full-cohort error. For \DMF{}, retaining the 20\% least-dispersed states reduced mean error by 12.2\%, while retaining 80\% still reduced it by 4.6\% (Fig.~\ref{fig:uq_cost}b). At 80\% retention, the corresponding reductions for the other generators were below 1\% or negligible. 
Fig.~\ref{fig:uq_cost}c examines the underlying state-wise association between spread and reconstruction difficulty. Across the same 200 states, the normalized ensemble spread of \DMF{} showed a Spearman correlation of $\rho=0.654$ with the relative-$L_2$ error of the ensemble-mean reconstruction, larger than the associations observed for the other generative models. 

The ablations further distinguish the role of the measurement-conditioning architecture from the generative transport itself. 
With the full model, the mean relative-$L_2$ error over the four unobserved fields was 0.1063 across 1,000 states. 
Removing the feedback from the global representation to the spatially anchored sensor tokens increased this error to 0.1270, while removing local query conditioning increased it to 0.1447 (Fig.~\ref{fig:uq_cost}d). Restricting conditioning to the local sensor pathway produced a much larger error of 0.3550.

The reference process affects a different aspect of reconstruction. Although the IID Gaussian source slightly reduced aggregate reconstruction error, it increased the mean high-frequency relative-\(L_2\) error of the streamwise velocity field from 0.873 to 1.825 compared with the smooth reference source (Fig.~\ref{fig:uq_cost}d,e).
The full model attenuates high-frequency $U_1$ energy relative to the truth. The IID source introduces excess power across much of the same range. 

We also measured the computational cost of transporting the physical field directly. At the native 40,300-point resolution, \DMF{} achieved an unobserved-field relative-$L_2$ error of 0.1063 with a median training-update time of 112.2~ms, peak training memory of 7.9~GiB and a median inference latency of 16.7~ms (Fig.~\ref{fig:uq_cost}f). Several deterministic or grid-based alternatives required less computation and had higher reconstruction error. 

\section*{Discussion}
\label{sec:Discussion}

Reconstruction from sparse sensors depends on two forms of structure. The model must combine information across observations and generate a spatially coherent field.
Functional flow matching provides a way to transport distributions of functions~\cite{Kerrigan2024FFM}. \DMF{} links this transport to a sensor--global--sensor representation. Its latent state summarizes measurements, and the transported state remains the physical field. Each query coordinate receives nearby evidence and field-wide context.

The results suggest that the apparently different reconstruction problems can share a common requirement: sparse measurements must influence the generated field without fixing its spatial support, output discretization, or observed physical channels.
The irregular-domain results show that sensor values can carry information about geometry that is not supplied explicitly.
The mixed-resolution results show that the learned reconstruction is not tied to the discretization of the complete fields used for training.
The combustion results further indicate that information can also be transferred across physical variables when entire target fields are unobserved.
These findings support a formulation in which sparse evidence is organized independently from the physical field being generated, while both global and spatially local measurement information remain available during transport.

The baseline comparisons and ablations provide further evidence that function-space transport alone does not explain the reconstruction gains.
FFM-FNO and FFM-Perceiver use functional transport with different backbones, yet remain less accurate than \DMF{} in the tested settings.
Within \DMF{}, removing global measurement exchange, feedback to sensor anchors or local query conditioning also degrades reconstruction in the combustion study.
These results point to the conditioning pathway as a key part of the formulation.

Future studies can test broader geometries, independent simulation trajectories, finer query grids, and different sensor layouts. The observation interface could also include spatial averages, line measurements, and other operators. Physical priors offer another direction when governing information is available~\cite{Wang2026FunDiff}. Larger three-dimensional fields may benefit from hierarchical sensor neighborhoods and domain-decomposed evaluation. These extensions would broaden the settings in which measurement-conditioned generation can recover useful physical fields.

\section*{Methods}
\label{sec:methods}

\subsection*{Field space, measurements and query evaluation}

Let $\mathcal{H}_{\Omega}$ be a separable Hilbert space supporting the law of a $C$-channel physical field. 
Sparse observations and requested outputs are represented separately by $\mathcal{Y}$ and $X$ in Eq.~\eqref{eq:results-measurements}. 
Point measurements and evaluations act on the field representatives supplied by the simulation and interpolation procedure. 

Shared tokenization and masked set attention make the conditioning representation invariant to sensor ordering.
Applying the same decoder at every query makes the output equivariant to query ordering. 
For unknown geometries, the fields and physical-support mask are represented on a common enclosing domain, and only the prescribed sensor measurements are provided as conditioning information. 

\subsection*{Gaussian reference process}

We approximate the Gaussian reference measure with a finite-rank random-Fourier-feature construction~\cite{Rahimi2007RFF}. 
For channel $c$,
\begin{equation}
U_{0,c}(\boldsymbol{x})
=\sqrt{\frac{2}{J}}
\sum_{j=1}^{J}
a_{cj}\cos(\boldsymbol{\omega}_j^{\top}\boldsymbol{x}+b_j),
\qquad a_{cj}\sim\mathcal{N}(0,1).
\label{eq:methods-rff}
\end{equation}
Frequencies $\boldsymbol{\omega}_j$ and phases $b_j$ are sampled once and retained as model buffers. The frequency scale sets the spatial correlation
length. 
Independent Gaussian coefficients are drawn for each field sample and channel. 
Conditional on the fixed feature basis, this construction defines a Gaussian law with positive semidefinite, finite-rank covariance. The IID Gaussian-value source is used in the prior ablation.

\subsection*{Conditional functional rectified flow}

Given a target field $U_1$ and its measurement set $\mathcal{Y}$, we independently sample $U_0$ and $\tau\sim\mathcal{U}(0,1)$ and form
\begin{equation}
X_{\tau}=(1-\tau)U_0+\tau U_1,
\qquad
V^{\star}(\tau,u;\mathcal{Y})
=\mathbb{E}[U_1-U_0\mid X_{\tau}=u,\mathcal{Y}].
\label{eq:methods-conditional-velocity}
\end{equation}
The Hilbert-space rectified-flow result connects this conditional expected velocity to the interpolation marginals under its stated rectifiability assumptions~\cite{Zhang2026FRF}. 
We approximate the velocity with the sensor-conditioned network and minimize the finite-query regression loss in Eq.~\eqref{eq:results-loss}. 
The implemented loss averages squared residuals over sampled queries and physical channels. 
Numerical integration of the learned velocity produces the reconstructed field.

\subsection*{Dataset preparation}
\label{subsec: Datasets}

\bmhead{Irregular-geometry datasets}
We adapted the elasticity and airfoil datasets released with Geo-FNO~\cite{Li2023GeoFNO} from forward-modeling benchmarks to sparse
reconstruction. 
The elasticity target contains the von Mises stress field and the material-support mask around an irregular hole. 
The airfoil target contains four compressible-flow fields (density, velocity components, and static pressure) and an airfoil/fluid-domain mask.
Prepared elasticity samples use Cartesian coordinates. 

\bmhead{Navier--Stokes multi-resolution dataset}
We retained only the streamwise velocity $V_x$ from the two-dimensional Navier--Stokes data in PDEBench~\cite{Takamoto2022PDEBench}. 
The selected cases were represented at $32\times32$,
$64\times64$ and $128\times128$ resolution and assigned to the five training recipes in Fig.~\ref{fig:mixed_resolution}a. 

\bmhead{Turbulent combustion}
We used an in-house large-eddy simulation of turbulent methane--air combustion over a two-dimensional backward-facing step. 
The simulation used \texttt{reactingFoam} in \texttt{OpenFOAM}~\cite{Jasak2009OpenFOAM}. For the present study, we retained five fields,
$Y_{\mathrm{CH_4}}$, $Y_{\mathrm{CO}}$, $T$, $U_1$ and $p$, at 40,300 grid points.

\subsection*{Baseline methods}
\label{subsec: BaselineMethods}

Each model receives the same task-specific measurement information and predicts the same fields. 

\bmhead{FFM-FNO}
FFM-FNO uses the Fourier neural operator parameterization introduced in functional flow matching~\cite{Kerrigan2024FFM,Li2021FNO}. 
We condition this velocity on sparse value/support maps concatenated with the evolving physical field and flow time. 

\bmhead{FFM-Perceiver}
FFM-Perceiver replaces the Fourier backbone with a Perceiver IO encoder--decoder~\cite{Kerrigan2024FFM,Jaegle2022PerceiverIO}.
The functional source and straight-flow objective are retained. 

\bmhead{SiT}
We adapt the scalable interpolant transformer~\cite{Ma2024SiT} to transport normalized physical-field arrays directly.

\bmhead{Latent FM}
Latent FM follows the two-stage construction of latent-space flow matching~\cite{Dao2023LatentFM}. 
An encoder $E$ and decoder $D$ first learn to reconstruct complete fields.

\bmhead{FNO}
The implemented supervised Fourier baseline maps sparse value/support maps directly to the target fields using FNO layers~\cite{Li2021FNO, Zhao2024RecFNO, Li2023GeoFNO}.
Its Cartesian implementation operates on the reconstruction grid. 
The irregular-coordinate wrapper scatters observations to a regular latent grid and bilinearly evaluates the predicted field at the requested coordinates. 

\bmhead{Senseiver}
Senseiver is the deterministic coordinate-query attention
baseline~\cite{Santos2023Senseiver}. 
\bmhead{MLP-RBF}
MLP-RBF is a non-attention reference constructed for this comparison~\cite{Chen_2023_ICCV}.
A pointwise regression head combines local evidence, query features, and a pooled global summary to predict the fields.

\bmhead{Author contributions}
L.W.: conceptualization, methodology, investigation, visualization, writing—original draft.
J.C.: methodology, investigation, visualization, writing—original draft.
N.T.: software, investigation, visualization.
Z.C.: methodology, investigation.
X.G.: methodology, writing—review and editing.
S.D.: supervision, writing—review and editing.

\bmhead{Data availability}

The two-dimensional CFD source data are available from the PDEBench repository (\url{https://github.com/pdebench/PDEBench}), and the elasticity and airfoil source data are available through the Geo-FNO repository (\url{https://github.com/neuraloperator/Geo-FNO}). 
The in-house turbulent-combustion dataset is large and is available from the corresponding authors upon reasonable request.

\bmhead{Code availability}

The DMF-Gen implementation, dataset interfaces, and scripts for training, evaluation, and reconstruction are available at \url{https://github.com/cosmos2w/DMF-Gen}.

\bmhead{Conflict of interest}
The authors declare no competing interests.

\clearpage
\begingroup
\small
\setlength{\bibsep}{0pt}
\bibliographystyle{unsrtnat}
\bibliography{references}

@article{Manohar2018SparseSensor,
  author  = {Manohar, Krithika and Brunton, Bingni W. and Kutz, J. Nathan and Brunton, Steven L.},
  title   = {Data-Driven Sparse Sensor Placement for Reconstruction: Demonstrating the Benefits of Exploiting Known Patterns},
  journal = {IEEE Control Systems Magazine},
  volume  = {38},
  number  = {3},
  pages   = {63--86},
  year    = {2018},
  doi     = {10.1109/MCS.2018.2810460}
}

@article{Fukami2021Voronoi,

  author = {Fukami, Kai and Maulik, Romit and Ramachandra, Nesar and Fukagata, Koji and Taira, Kunihiko},
  title   = {Global field reconstruction from sparse sensors with {Voronoi} tessellation-assisted deep learning},
  journal = {Nature Machine Intelligence},
  volume  = {3},
  pages   = {945--951},
  year    = {2021},
  doi     = {10.1038/s42256-021-00402-2},
  number = {11}
}

@article{Santos2023Senseiver,
  author  = {Santos, Javier E. and Fox, Zachary R. and Mohan, Arvind and O'Malley, Daniel and Viswanathan, Hari and Lubbers, Nicholas},
  title   = {Development of the {Senseiver} for efficient field reconstruction from sparse observations},
  journal = {Nature Machine Intelligence},
  volume  = {5},
  pages   = {1317--1325},
  year    = {2023},
  doi     = {10.1038/s42256-023-00746-x}
}

@article{Callaham2019Sparse,
  author  = {Callaham, Jared L. and Maeda, Kazuki and Brunton, Steven L.},
  title   = {Robust flow reconstruction from limited measurements via sparse representation},
  journal = {Physical Review Fluids},
  volume  = {4},
  pages   = {103907},
  year    = {2019},
  doi     = {10.1103/PhysRevFluids.4.103907}
}

@article{Wang2024ChemFlow,
  author  = {Wang, Linzheng and Deng, Ruiqu and Zhang, Ruizhi and Luo, Yonghao and Deng, Sili},
  title   = {3-D full-field reconstruction of chemically reacting flow towards high-dimension conditions through machine learning},
  journal = {Chemical Engineering Journal},
  volume  = {499},
  pages   = {156435},
  year    = {2024},
  doi     = {10.1016/j.cej.2024.156435}
}

@article{Gou2024Global,
  title = {Global high-resolution total water storage anomalies from self-supervised data assimilation using deep learning algorithms},
  volume = {2},
  url = {https://doi.org/10.1038/s44221-024-00194-w},
  DOI = {10.1038/s44221-024-00194-w},
  number = {2},
  journal = {Nature Water},
  author = {Gou,  Junyang and Soja,  Benedikt},
  year = {2024},
  pages = {139--150}
}

@article{Lenain2026unprecedented,
  title = {An unprecedented view of ocean currents from geostationary satellites},
  volume = {19},
  url = {https://doi.org/10.1038/s41561-026-01943-0},
  DOI = {10.1038/s41561-026-01943-0},
  number = {5},
  journal = {Nature Geoscience},
  author = {Lenain,  Luc and Srinivasan,  Kaushik and Barkan,  Roy and Pizzo,  Nick},
  year = {2026},
  pages = {526--533}
}

@article{Wang2026DigitalTwin,
  author  = {Wang, Linzheng and Li, Yaojun and Deng, Sili},
  title   = {Building Digital Twin for {3D} Multi-Field Reconstruction and Optimization of Industrial-Scale Combustion Systems},
  journal = {Engineering},
  volume  = {59},
  pages   = {197--214},
  year    = {2026},
  doi     = {10.1016/j.eng.2025.08.020}
}

@article{Willcox2006Gappy,
  author  = {Willcox, Karen},
  title   = {Unsteady flow sensing and estimation via the gappy proper orthogonal decomposition},
  journal = {Computers \& Fluids},
  volume  = {35},
  number  = {2},
  pages   = {208--226},
  year    = {2006},
  doi     = {10.1016/j.compfluid.2004.11.006}
}

@article{Brunton2016Placement,
  title = {Sparse Sensor Placement Optimization for Classification},
  volume = {76},
  url = {https://doi.org/10.1137/15M1036713},
  DOI = {10.1137/15m1036713},
  number = {5},
  journal = {SIAM Journal on Applied Mathematics},
  author = {Brunton,  B. W. and Brunton,  S. L. and Proctor,  J. L. and Kutz,  J. N.},
  year = {2016},
  pages = {2099--2122}
}

@article{Achermann2024WindSeer,
  title = {{WindSeer}: real-time volumetric wind prediction over complex terrain aboard a small uncrewed aerial vehicle},
  volume = {15},
  url = {https://doi.org/10.1038/s41467-024-47778-4},
  DOI = {10.1038/s41467-024-47778-4},
  number = {1},
  journal = {Nature Communications},
  author = {Achermann,  Florian and Stastny,  Thomas and Danciu,  Bogdan and Kolobov,  Andrey and Chung,  Jen Jen and Siegwart,  Roland and Lawrance,  Nicholas},
  year = {2024},
  pages = {3507}
}

@article{Li2024S3GM,
  author  = {Li, Zeyu and Han, Wang and Zhang, Yue and Fu, Qingfei and Li, Jingxuan and Qin, Lizi and Dong, Ruoyu and Sun, Hao and Deng, Yue and Yang, Lijun},
  title   = {Learning spatiotemporal dynamics with a pretrained generative model},
  journal = {Nature Machine Intelligence},
  volume  = {6},
  number  = {12},
  pages   = {1566--1579},
  year    = {2024},
  doi     = {10.1038/s42256-024-00938-z}
}

@article{Wang2026UnifiedManifold,
  author  = {Wang, Linzheng and Chen, Zituo and Li, Yaojun and Deng, Ruiqu and Zhang, Ruizhi and Luo, Yonghao and Lu, Lu and Deng, Sili},
  title   = {Learning multi-physical system on a unified manifold by collaboratively fused features},
  journal = {Engineering Applications of Artificial Intelligence},
  volume  = {170},
  pages   = {114178},
  year    = {2026},
  doi     = {10.1016/j.engappai.2026.114178}
}

@article{Liu2025Enhancing,
  title = {Enhancing deep learning-based field reconstruction with a differentiable learning framework},
  volume = {7},
  url = {https://doi.org/10.1038/s42256-025-01063-1},
  DOI = {10.1038/s42256-025-01063-1},
  number = {7},
  journal = {Nature Machine Intelligence},
  author = {Liu,  Xu and Peng,  Wei and Zhang,  Xiaoya and Zhao,  Xiaoyu and Zhou,  Weien and Yao,  Wen and Chen,  Xiaoqian},
  year = {2025},
  pages = {1129--1140}
}

@article{Li2023GeoFNO,
  author  = {Li, Zongyi and Huang, Daniel Zhengyu and Liu, Burigede and Anandkumar, Anima},
  title   = {{Fourier} Neural Operator with Learned Deformations for {PDEs} on General Geometries},
  journal = {Journal of Machine Learning Research},
  volume  = {24},
  number  = {388},
  pages   = {1--26},
  year    = {2023},
  url     = {https://www.jmlr.org/papers/v24/23-0064.html}
}

@article{Zhao2024RecFNO,
  title = {{RecFNO}: A resolution-invariant flow and heat field reconstruction method from sparse observations via {Fourier} neural operator},
  volume = {195},
  url = {https://doi.org/10.1016/j.ijthermalsci.2023.108619},
  DOI = {10.1016/j.ijthermalsci.2023.108619},
  journal = {International Journal of Thermal Sciences},
  author = {Zhao,  Xiaoyu and Chen,  Xiaoqian and Gong,  Zhiqiang and Zhou,  Weien and Yao,  Wen and Zhang,  Yunyang},
  year = {2024},
  pages = {108619}
}

@article{Wang2026Developing,
  title = {Developing multi-field sparse reconstruction digital twins: A study in ammonia combustor},
  volume = {42},
  url = {https://doi.org/10.1016/j.proci.2026.106421},
  DOI = {10.1016/j.proci.2026.106421},
  journal = {Proceedings of the Combustion Institute},
  author = {Wang,  Linzheng and Chen,  Jason and Piscopo,  Alessandro and Parente,  Alessandro and Deng,  Sili},
  year = {2026},
  pages = {106421}
}

@article{Wright2009Sparse,
  title = {Sparse Reconstruction by Separable Approximation},
  volume = {57},
  url = {https://doi.org/10.1109/TSP.2009.2016892},
  DOI = {10.1109/tsp.2009.2016892},
  number = {7},
  journal = {IEEE Transactions on Signal Processing},
  author = {Wright,  S.J. and Nowak,  R.D. and Figueiredo,  M.A.T.},
  year = {2009},
  pages = {2479--2493}
}

@inproceedings{Yao2025FunDPS,
  author    = {Yao, Jiachen and Mammadov, Abbas and Berner, Julius and Kerrigan, Gavin and Ye, Jong Chul and Azizzadenesheli, Kamyar and Anandkumar, Anima},
  title     = {Guided Diffusion Sampling on Function Spaces with Applications to {PDEs}},
  booktitle = {Advances in Neural Information Processing Systems},
  volume    = {38},

  pages = {127057--127094},
  year      = {2025},
  doi       = {10.52202/085713-4233}
}

@inproceedings{
mohseni2025spectral,
title={Spectral Convolutional Conditional Neural Processes},
author={Peiman Mohseni and Nick Duffield},

  booktitle = {Advances in Neural Information Processing Systems},
year={2025},

  url = {https://papers.neurips.cc/paper_files/paper/2025/hash/39a5e95b048d8184d530bd0c81afa46a-Abstract-Conference.html},
  volume = {38},
  pages = {40416--40446},
  doi = {10.52202/085713-1350}
}

@article{Lu2021DeepONet,
  author  = {Lu, Lu and Jin, Pengzhan and Pang, Guofei and Zhang, Zhongqiang and Karniadakis, George Em},
  title   = {Learning nonlinear operators via {DeepONet} based on the universal approximation theorem of operators},
  journal = {Nature Machine Intelligence},
  volume  = {3},
  pages   = {218--229},
  year    = {2021},
  doi     = {10.1038/s42256-021-00302-5}
}

@inproceedings{Li2021FNO,
  author    = {Li, Zongyi and Kovachki, Nikola and Azizzadenesheli, Kamyar and Liu, Burigede and Bhattacharya, Kaushik and Stuart, Andrew and Anandkumar, Anima},
  title     = {{Fourier} Neural Operator for Parametric Partial Differential Equations},
  booktitle = {International Conference on Learning Representations},
  year      = {2021},
  url       = {https://openreview.net/forum?id=c8P9NQVtmnO}
}

@article{Kovachki2023NeuralOperator,
  author  = {Kovachki, Nikola and Li, Zongyi and Liu, Burigede and Azizzadenesheli, Kamyar and Bhattacharya, Kaushik and Stuart, Andrew and Anandkumar, Anima},
  title   = {Neural Operator: Learning Maps Between Function Spaces With Applications to {PDEs}},
  journal = {Journal of Machine Learning Research},
  volume  = {24},
  number  = {89},
  pages   = {1--97},
  year    = {2023},
  url     = {https://www.jmlr.org/papers/v24/21-1524.html}
}

@article{Azizzadenesheli2024NeuralOperators,
  author  = {Azizzadenesheli, Kamyar and Kovachki, Nikola and Li, Zongyi and Liu-Schiaffini, Miguel and Kossaifi, Jean and Anandkumar, Anima},
  title   = {Neural operators for accelerating scientific simulations and design},
  journal = {Nature Reviews Physics},
  volume  = {6},
  pages   = {320--328},
  year    = {2024},
  doi     = {10.1038/s42254-024-00712-5}
}

@inproceedings{Huang2024DiffusionPDE,
  author    = {Huang, Jiahe and Yang, Guandao and Wang, Zichen and Park, Jeong Joon},
  title     = {{DiffusionPDE}: Generative PDE-Solving under Partial Observation},
  booktitle = {Advances in Neural Information Processing Systems},
  volume    = {37},
  pages     = {130291--130323},
  year      = {2024},
  url       = {https://proceedings.neurips.cc/paper_files/paper/2024/hash/eb3878c1dcbfff9ee95d5d033e5f5942-Abstract-Conference.html},
  doi = {10.52202/079017-4140}
}

@article{Du2024CoNFiLD,
  author  = {Du, Pan and Parikh, Meet Hemant and Fan, Xiantao and Liu, Xin-Yang and Wang, Jian-Xun},
  title   = {Conditional neural field latent diffusion model for generating spatiotemporal turbulence},
  journal = {Nature Communications},
  volume  = {15},
  pages   = {10416},
  year    = {2024},
  doi     = {10.1038/s41467-024-54712-1}
}

@article{Wang2026FunDiff,
  author  = {Wang, Sifan and Dou, Zehao and Shan, Siming and Liu, Tong-Rui and Lu, Lu},
  title   = {{FunDiff}: diffusion models over function spaces for physics-informed generative modeling},
  journal = {Nature Communications},
  volume  = {17},
  pages   = {5749},
  year    = {2026},
  doi     = {10.1038/s41467-026-72292-0}
}

@inproceedings{Rombach2022LDM,
  author    = {Rombach, Robin and Blattmann, Andreas and Lorenz, Dominik and Esser, Patrick and Ommer, Bjorn},
  title     = {High-Resolution Image Synthesis with Latent Diffusion Models},
  booktitle = {Proceedings of the IEEE/CVF Conference on Computer Vision and Pattern Recognition},
  pages     = {10684--10695},
  year      = {2022},
  doi       = {10.1109/CVPR52688.2022.01042}
}

@inproceedings{Hoogeboom2023SimpleDiffusion,
  author    = {Hoogeboom, Emiel and Heek, Jonathan and Salimans, Tim},
  title     = {Simple diffusion: End-to-end diffusion for high resolution images},
  booktitle = {Proceedings of the 40th International Conference on Machine Learning},
  series    = {Proceedings of Machine Learning Research},
  volume    = {202},
  pages     = {13213--13232},
  year      = {2023},
  url       = {https://proceedings.mlr.press/v202/hoogeboom23a.html}
}

@InProceedings{Chen_2026_CVPR,
    author    = {Chen, Zhennan and Zhu, Junwei and Chen, Xu and Zhang, Jiangning and Hu, Xiaobin and Zhao, Hanzhen and Wang, Chengjie and Yang, Jian and Tai, Ying},
    title     = {{DiP}: Taming Diffusion Models in Pixel Space},
    booktitle = {Proceedings of the IEEE/CVF Conference on Computer Vision and Pattern Recognition (CVPR)},
    year      = {2026},
    pages     = {36136-36146},
  url = {https://openaccess.thecvf.com/content/CVPR2026/html/Chen_DiP_Taming_Diffusion_Models_in_Pixel_Space_CVPR_2026_paper.html}
}

@inproceedings{ICLR2026_47f8478e,
 author = {Wang, Shuai and Gao, Ziteng and Zhu, Chenhui and Huang, Weilin and Wang, Limin},
 booktitle = {International Conference on Learning Representations},
 pages = {43559--43580},
 title = {{PixNerd}: Pixel Neural Field Diffusion},
 url = {https://proceedings.iclr.cc/paper_files/paper/2026/file/47f8478e53a40b9ba830b113fea01fdf-Paper-Conference.pdf},
 volume = {2026},
 year = {2026}
}

@InProceedings{Li_2026_CVPR,
    author    = {Li, Tianhong and He, Kaiming},
    title     = {Back to Basics: Let Denoising Generative Models Denoise},
    booktitle = {Proceedings of the IEEE/CVF Conference on Computer Vision and Pattern Recognition (CVPR)},
    year      = {2026},
    pages     = {36115-36125},
  url = {https://openaccess.thecvf.com/CVPR2026?day=2026-06-07}
}

@inproceedings{Kerrigan2024FFM,
  author    = {Kerrigan, Gavin and Migliorini, Giosue and Smyth, Padhraic},
  title     = {Functional Flow Matching},
  booktitle = {Proceedings of the 27th International Conference on Artificial Intelligence and Statistics},
  series    = {Proceedings of Machine Learning Research},
  volume    = {238},
  pages     = {3934--3942},
  year      = {2024},
  url       = {https://proceedings.mlr.press/v238/kerrigan24a.html}
}

@article{Lim2025FunctionSpaceDiffusion,
  author  = {Lim, Jae Hyun and Kovachki, Nikola B. and Baptista, Ricardo and Beckham, Christopher and Azizzadenesheli, Kamyar and Kossaifi, Jean and Voleti, Vikram and Song, Jiaming and Kreis, Karsten and Kautz, Jan and Pal, Christopher and Vahdat, Arash and Anandkumar, Anima},
  title   = {Score-Based Diffusion Models in Function Space},
  journal = {Journal of Machine Learning Research},
  volume  = {26},
  number  = {158},
  pages   = {1--62},
  year    = {2025},
  url     = {https://www.jmlr.org/papers/v26/23-1472.html}
}

@inproceedings{Zhang2026FRF,
  author    = {Zhang, Jianxin and Scott, Clayton},
  title     = {Flow Straight and Fast in {Hilbert} Space: Functional Rectified Flow},
  booktitle = {International Conference on Learning Representations},
  year      = {2026},

  url = {https://proceedings.iclr.cc/paper_files/paper/2026/hash/1704ddd0bb89f159dfe609b32c889995-Abstract-Conference.html},
  eprint    = {2509.10384},
  archivePrefix = {arXiv},
  pages = {13612--13649}
}

@inproceedings{Liu2023RectifiedFlow,
  author    = {Liu, Xingchao and Gong, Chengyue and Liu, Qiang},
  title     = {Flow Straight and Fast: Learning to Generate and Transfer Data with Rectified Flow},
  booktitle = {International Conference on Learning Representations},
  year      = {2023},
  url       = {https://openreview.net/forum?id=XVjTT1nw5z}
}

@article{Berner2026Principled,
  author  = {Berner, Julius and Liu-Schiaffini, Miguel and Kossaifi, Jean and Duruisseaux, Valentin and Bonev, Boris and Azizzadenesheli, Kamyar and Anandkumar, Anima},
  title   = {Principled approaches for extending neural architectures to function spaces for operator learning},
  journal = {Nature Machine Intelligence},
  volume  = {8},
  pages   = {1173--1181},
  year    = {2026},
  doi     = {10.1038/s42256-026-01267-z}
}

@article{Calvello2025ContinuumAttention,
  author  = {Edoardo Calvello and Nikola B. Kovachki and Matthew E. Levine and Andrew M. Stuart},
  title   = {Continuum Attention for Neural Operators},
  journal = {Journal of Machine Learning Research},
  year    = {2025},
  volume  = {26},
  number  = {300},
  pages   = {1--52},
  url     = {http://jmlr.org/papers/v26/24-0879.html}
}

@misc{Wang2026GLU,
  author = {Wang, Linzheng and Chen, Jason and Tricard, Nicolas and Chen, Zituo and Deng, Sili},
  title = {{GLU}: Global-Local-Uncertainty Fusion for Scalable Spatiotemporal Reconstruction and Forecasting},
  year = {2026},
  eprint = {2603.26023},
  archivePrefix = {arXiv},
  primaryClass = {cs.LG},
  doi = {10.48550/arXiv.2603.26023},
  url = {https://arxiv.org/abs/2603.26023}
}

@article{Shi2025MINO,
  author  = {Shi, Yaozhong and Ross, Zachary E. and Asimaki, Domniki and Azizzadenesheli, Kamyar},
  title   = {Mesh-Informed Neural Operator: A Transformer Generative Approach},
  journal = {Transactions on Machine Learning Research},
  year    = {2025},
  url     = {https://openreview.net/forum?id=K8qAuRfv0G}
}

@inproceedings{Takamoto2022PDEBench,
  author    = {Takamoto, Makoto and Praditia, Timothy and Leiteritz, Raphael and MacKinlay, Daniel and Alesiani, Francesco and Pfluger, Dirk and Niepert, Mathias},
  title     = {{PDEBench}: An Extensive Benchmark for Scientific Machine Learning},
  booktitle = {Advances in Neural Information Processing Systems},
  volume    = {35},
  pages     = {1596--1611},
  year      = {2022},
  url = {https://proceedings.neurips.cc/paper_files/paper/2022/hash/0a9747136d411fb83f0cf81820d44afb-Abstract-Datasets_and_Benchmarks.html}
}

@inproceedings{Jaegle2022PerceiverIO,
  author = {Jaegle, Andrew and Borgeaud, Sebastian and Alayrac, Jean-Baptiste and Doersch, Carl and Ionescu, Catalin and Ding, David and Koppula, Skanda and Zoran, Daniel and Brock, Andrew and Shelhamer, Evan and H{\'e}naff, Olivier and Botvinick, Matthew M. and Zisserman, Andrew and Vinyals, Oriol and Carreira, Jo{\~a}o},
  title = {{Perceiver IO}: A General Architecture for Structured Inputs \& Outputs},
  booktitle = {International Conference on Learning Representations},
  year = {2022},
  url = {https://openreview.net/forum?id=fILj7WpI-g}
}

@inproceedings{Ma2024SiT,
  author    = {Ma, Nanye and Goldstein, Mark and Albergo, Michael S. and Boffi, Nicholas M. and Vanden-Eijnden, Eric and Xie, Saining},
  title     = {{SiT}: Exploring Flow and Diffusion-Based Generative Models with Scalable Interpolant Transformers},
  booktitle = {Computer Vision -- ECCV 2024},
  series    = {Lecture Notes in Computer Science},
  volume    = {15135},
  pages     = {23--40},
  year      = {2024},
  address   = {Cham},
  doi       = {10.1007/978-3-031-72980-5_2},
  isbn      = {978-3-031-72980-5}
}

@misc{Dao2023LatentFM,
  author = {Dao, Quan and Phung, Hao and Nguyen, Binh and Tran, Anh},
  title = {Flow Matching in Latent Space},
  year = {2023},
  eprint = {2307.08698},
  archivePrefix = {arXiv},
  primaryClass = {cs.CV},
  doi = {10.48550/arXiv.2307.08698},
  url = {https://arxiv.org/abs/2307.08698}
}

@article{Xue2014GMSD,
  author = {Xue, Wufeng and Zhang, Lei and Mou, Xuanqin and Bovik, Alan C.},
  title = {Gradient Magnitude Similarity Deviation: A Highly Efficient Perceptual Image Quality Index},
  journal = {IEEE Transactions on Image Processing},
  volume = {23},
  number = {2},
  pages = {684--695},
  year = {2014},
  doi = {10.1109/TIP.2013.2293423}
}

@article{Gomes2025sparse,
  title = {A sparse and decoupled supervision framework for multi-field reconstruction in turbulent flows using physics-informed neural networks},
  volume = {37},
  url = {https://doi.org/10.1063/5.0269730},
  DOI = {10.1063/5.0269730},
  number = {7},
  journal = {Physics of Fluids},
  author = {Gomes,  David and Espírito Santo,  António and Páscoa,  José C.},
  year = {2025},
  pages = {075113}
}

@article{Kobayashi2026VirtualSensing,
  title   = {Virtual Sensing to Enable Real-Time Monitoring of Inaccessible Locations \& Unmeasurable Parameters},
  author  = {Kobayashi, Kazuma and Ahmed, Farid and Park, Jaewan and
             Sarkar, Subhankar and Koric, Seid and Chakraborty, Souvik and
             Alam, Syed Bahauddin},
  journal = {Nature Communications},
  year    = {2026},
  doi     = {10.1038/s41467-026-77463-7},

  volume = {17},
  number = {1},
  pages = {10976}
}

@inproceedings{Rahimi2007RFF,
  author = {Rahimi, Ali and Recht, Benjamin},
  title = {Random Features for Large-Scale Kernel Machines},
  booktitle = {Advances in Neural Information Processing Systems},
  volume = {20},
  year = {2007},
  url = {https://proceedings.neurips.cc/paper_files/paper/2007/hash/013a006f03dbc5392effeb8f18fda755-Abstract.html},
  pages = {1177--1184}
}

@article{Jasak2009OpenFOAM,
  author = {Jasak, Hrvoje},
  title = {{OpenFOAM}: Open source {CFD} in research and industry},
  journal = {International Journal of Naval Architecture and Ocean Engineering},
  volume = {1},
  number = {2},
  pages = {89--94},
  year = {2009},
  doi = {10.2478/IJNAOE-2013-0011}
}

@InProceedings{Chen_2023_ICCV,
    author    = {Chen, Zhang and Li, Zhong and Song, Liangchen and Chen, Lele and Yu, Jingyi and Yuan, Junsong and Xu, Yi},
    title     = {{NeuRBF}: A Neural Fields Representation with Adaptive Radial Basis Functions},
    booktitle = {Proceedings of the IEEE/CVF International Conference on Computer Vision (ICCV)},
    year      = {2023},
    pages     = {4182-4194},
  url = {https://openaccess.thecvf.com/content/ICCV2023/html/Chen_NeuRBF_A_Neural_Fields_Representation_with_Adaptive_Radial_Basis_Functions_ICCV_2023_paper.html}
}
\endgroup
\end{document}